\pdfoutput=1

\documentclass[11pt]{article}

\usepackage{ACL2023}

\usepackage{times}
\usepackage{latexsym}

\usepackage[T1]{fontenc}

\usepackage[utf8]{inputenc}

\usepackage{microtype}

\usepackage{inconsolata}

\usepackage{xspace,mfirstuc,tabulary}
\usepackage[section]{algorithm}
\usepackage[noend]{algpseudocode}
\usepackage{relsize}
\usepackage{graphicx}
\usepackage{caption}
\usepackage{subcaption}
\usepackage{amsmath}
\usepackage{bm}
\usepackage{amsfonts}
\usepackage{booktabs}
\usepackage{multirow}
\usepackage{mathtools}
\usepackage{siunitx}
\mathtoolsset{showonlyrefs=true}
\newcommand{\innerproduct}[2]{\left\langle #1, #2 \right\rangle}
\newcommand\identity{1\kern-0.25em\text{l}}

\def\Figref#1{Figure~\ref{#1}}

\def\Secref#1{\S\ref{#1}}

\def\eqref#1{equation~\ref{#1}}

\def\1{\bm{1}}

\def\va{{\bm{a}}}
\def\vb{{\bm{b}}}

\def\vp{{\bm{p}}}

\def\vs{{\bm{s}}}
\def\vt{{\bm{t}}}

\def\evp{{p}}

\def\mC{{\bm{C}}}

\def\mP{{\bm{P}}}

\def\mU{{\bm{U}}}

\def\mW{{\bm{W}}}

\def\mY{{\bm{Y}}}

\DeclareMathAlphabet{\mathsfit}{\encodingdefault}{\sfdefault}{m}{sl}
\SetMathAlphabet{\mathsfit}{bold}{\encodingdefault}{\sfdefault}{bx}{n}

\def\sY{{\mathbb{Y}}}

\def\emC{{C}}

\def\emP{{P}}

\def\emY{{Y}}

\newcommand{\R}{\mathbb{R}}


\newcommand{\Tref}[1]{Table~\ref{#1}}
\newcommand{\Appendixref}[1]{Appendix~\ref{#1}}

\title{Unbalanced Optimal Transport for Unbalanced Word Alignment}

\author{Yuki Arase \\
  Osaka University, Japan \\
  \texttt{arase@ist.osaka-u.ac.jp} \\\And
  Han Bao \\
  Kyoto University, Japan \\
  \texttt{bao@i.kyoto-u.ac.jp} \\\And
  Sho Yokoi \\
  Tohoku University, Japan \\
  RIKEN, Japan \\
  \texttt{yokoi@tohoku.ac.jp} \\
  }

\begin{document}
\maketitle
\begin{abstract}
Monolingual word alignment is crucial to model semantic interactions between sentences.
In particular, null alignment, a phenomenon in which words have no corresponding counterparts, is pervasive and critical in handling semantically divergent sentences. 
Identification of null alignment is useful on its own to reason about the semantic similarity of sentences by indicating there exists information inequality. 
To achieve \emph{unbalanced} word alignment that values both alignment and null alignment, this study shows that the family of optimal transport (OT), i.e., balanced, partial, and unbalanced OT, are natural and powerful approaches even without tailor-made techniques.
Our extensive experiments covering unsupervised and supervised settings indicate that our generic OT-based alignment methods are competitive against the state-of-the-arts specially designed for word alignment, remarkably on challenging datasets with high null alignment frequencies. 
\end{abstract}

\section{Introduction} 
Monolingual word alignment, which identifies semantically corresponding words in a sentence pair, has been actively studied as a crucial technique for modelling semantic relationships between sentences, such as for paraphrase identification, textual entailment recognition, and question answering \cite{maccartney:08,das-smith:2009:ACLIJCNLP,wang-manning-2010-probabilistic,heilman-smith:2010:NAACLHLT2,yao:13a,feldman-el-yaniv-2019-multi}. 
Its ability to declare redundant information in sentences is also useful for summarisation and sentence fusion \cite{thadani-mckeown-2013-supervised,brook-weiss-etal-2021-qa}. 
In addition, the alignment information is valuable for interpretability of model predictions \cite{agirre-etal-2015-semeval,li-srikumar-2016-exploiting} and for realising interactive document exploration as well \cite{shapira-etal-2017-interactive,hirsch-etal-2021-ifacetsum}.

\begin{figure}[t!]
\centering
\includegraphics[width=0.9\linewidth]{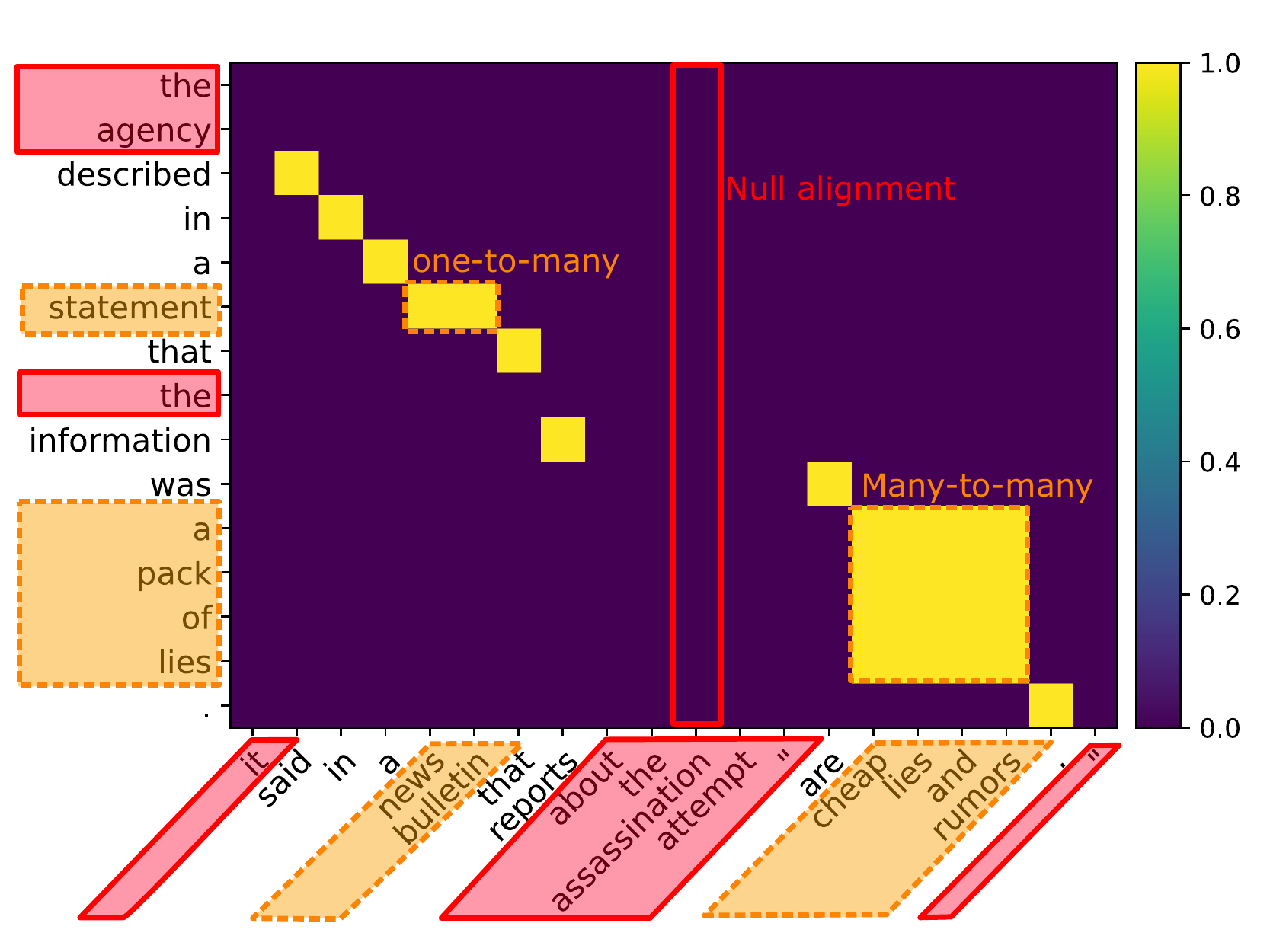}
\caption{Unbalanced monolingual word alignment matrix; frequent null alignment (enclosed in \textcolor{red}{red boxes}) and mapping beyond one-to-one (enclosed in \textcolor{orange}{orange dashed boxes}) are primary challenges.}
\label{fig:gold_alignment}
\end{figure}

\Figref{fig:gold_alignment} illustrates the challenges of monolingual word alignment.  
The first challenge is the \emph{null alignment}, where words may not have corresponding counterparts, which causes alignment asymmetricity \cite{maccartney:08}. 
Null alignment is prevalent in semantically divergent sentences; indeed, the null alignment ratio reaches $63.8\%$ in entailment sentence pairs used in our experiments (shown in the third row of \Tref{tab:main_unsupervised_results_f1}).  
Its identification explicitly declares semantic gaps between two sentences and helps reason about their semantic (dis)similarity \cite{li-srikumar-2016-exploiting}.
The second challenge is that alignment \emph{beyond one-to-one mapping} needs to be addressed. 
These challenges constitute an \emph{unbalanced} word alignment problem, where both word-by-word and null alignment should be fully identified.

This study reveals that a family of optimal transport (OT) are suitable tools for unbalanced word alignment. 
Among the OT problems, balanced OT (\textbf{BOT}) \cite{monge1781memoire,kantorovich-1942}\footnote{We explicitly call it \emph{balanced} OT (BOT) to distinguish it from partial and unbalanced OT in this paper.} should be the most prominent in natural language processing (NLP) \cite{wan20073718,wmd}, which can handle many-to-many alignment. 
In contrast to BOT that is unable to deal with null alignment, partial OT (\textbf{POT}) \cite{caffarelli-et-al-2010,figalli-2010} and unbalanced OT (\textbf{UOT}) \cite{cortes-et-al-2015,chizat-et-al-2016} can handle the asymmetricity as desired in unbalanced word alignment, which has also attracted applications where null alignment is unignorable \cite{swanson-etal-2020-rationalizing,zhang-etal-2020-semantic-matching,yu-etal-2022-optimal,chen-et-al-ijcai2020,pmlr-v129-wang20c}.

This is the first study that connects these two paradigms of unbalanced word alignment and the family of OT problems that can naturally address null alignment as well as many-to-many alignment.  
We empirically (1)~demonstrate that the OT-based methods are natural and sufficiently powerful approaches to unbalanced word alignment without tailor-made techniques and (2)~deliver a comprehensive picture that unveils the characteristics of BOT, POT, and UOT on unbalanced word alignment with different null alignment ratios. 
We conduct extensive experiments using representative datasets for understanding OT-based alignment: effects of OT formalisation, regularisation, and a heuristic to sparsify alignments. 
Our primary findings can be summarised as follows. 
First, in unsupervised alignment, the best OT problem depends on null alignment ratios. 
Second, simple thresholding on regularised BOT can produce unbalanced alignment. 
Third, in supervised alignment, simple and generic OT-based alignment shows competitive performance to the state-of-the-art models specially designed for word alignment. 

The adoption of well-understood methods with a solid theory like OT is highly valuable in application development and scientific reproducibility. 
Furthermore, OT-based alignment performs superiorly or competitively to existing methods, despite being independent of tailor-made techniques. 
Our OT-based alignment methods are publicly available as a tool called \emph{OTAlign}.\footnote{\url{https://github.com/yukiar/OTAlign}}

\section{Related Work}
\subsection{Word Alignment Problem}
\label{sec:related_worK_wa}
Word alignment techniques have been actively studied in the crosslingual setting, primarily for machine translation. 
Crosslingual and monolingual word alignment tackle relevant problems; however, they have notable differences \cite{maccartney:08}. 
Crosslingual alignment can assume that a large-scale parallel corpus exists, which is scarce in the monolingual case. 
Furthermore, asymmetricity can be more intense in monolingual alignment to handle semantically divergent sentences. 
A common approach to crosslingual word alignment is unsupervised learning using parallel corpora \cite{och-ney-2003-systematic,fastalign,garg-etal-2019-jointly,zenkel-etal-2020-end,dou-neubig-2021-word}. 
Among them, \citet{dou-neubig-2021-word} applied BOT for alignment while focussing on fine-tuning multilingual pre-trained language models using a parallel corpus. 
Among crosslingual word alignment methods, SimAlign \cite{simalign} is directly applicable to monolingual alignment because it only uses a multilingual pre-trained language model and simple heuristics without parallel corpora. In addition, the supervised word alignment method proposed by \citet{nagata-etal-2020-supervised}, which modelled alignment as a question-answering task, is also applicable. 

For monolingual word alignment, supervised learning has been commonly used \cite{maccartney:08,thadani:11,thadani:12}. 
A representative method modelled word alignment using the conditional random field regarding source words as observations and target words as hidden states \cite{yao:13a,yao-etal-2013-semi}. 
\citet{lan-etal-2021-neural} enhanced this approach by adopting neural models, which constitute the state-of-the-art methods together with \citet{nagata-etal-2020-supervised}. 
In monolingual alignment, phrase alignment is also a research focus \cite{ouyang-mckeown-2019-neural,arase:emnlp2020,culkin-etal-2021-iterative}, which depends on high-quality parsers and chunkers. 
In contrast, we use words as an alignment unit and do not assume the availability of such parsers.

\subsection{Optimal Transport in NLP}
\label{sec:related_work_ot4nlp}
BOT, POT, and UOT have been adopted in various NLP tasks where alignment exists implicitly or explicitly.  
Most previous studies modelled tasks of their interest, with alignment being a hidden state; further, the alignment quality itself was out of their focus. 
A typical application is similarity estimation among sentences and documents using BOT \cite{wmd,gao-et-al-NIPS2016,zhao-etal-2019-moverscore,yokoi-etal-2020-word,chen-et-al-ijcai2020,alqahtani-etal-2021-using-optimal,lee-etal-2022-toward,mysore-etal-2022-multi,wang-etal-2022-unsupervised} and recently UOT \cite{chen-et-al-ijcai2020,pmlr-v129-wang20c}. 
Such similarity estimation mechanisms can be integrated into language generation models as a penalty \cite{chen2018improving,zhang-etal-2020-topic,li-etal-2020-improving-text}.

Unlike the tasks above, alignment information obtained with BOT, POT, and UOT is the primary concern in extractive summarization \cite{tang-etal-2022-otextsum}, text matching \cite{swanson-etal-2020-rationalizing,zhang-etal-2020-semantic-matching,yu-etal-2022-optimal}, and bilingual lexicon induction \cite{zhang-etal-2017-earth,grave19a,zhao-etal-2020-relaxed}. 
However, these tasks and unbalanced monolingual word alignment have different objectives. These tasks concern the precision where \emph{alignment exists}. 
In contrast, unbalanced word alignment aims at exhaustive identification of both alignment and null alignment. 
Furthermore, there has not been any systematic study that compared the qualities of different OT-based alignment methods, which hinders the understanding of a suitable OT formulation for problems with different null alignment frequencies.

\section{Problem Definition}
\label{sec:problem_definition}
Suppose we have a source and target sentence pair $s = \{s_1, s_2,\dots,s_n\}$ and $t = \{t_1, t_2, \dots, t_m\}$ with their word embeddings $\{\vs_1, \vs_2, \cdots, \vs_n\}$ and $\{\vt_1, \vt_2, \cdots, \vt_m\}$, respectively, where $\vs_i, \vt_j \in \R^d$. 
The goal of monolingual word alignment is to identify an alignment $\mP \in \R_+^{n\times m}$ between semantically corresponding words, where $\emP_{i,j}$ indicates a likelihood or binary indicator of aligning $s_i$ and $t_j$. 

\paragraph{Evaluation Metrics for Unbalanced Alignment}
As evaluation metrics, we adopt the macro averages of precision and recall of alignment pairs, and their F$1$ score \cite{maccartney:08}. 
In light of the importance of null alignment, we explicitly incorporate it into the evaluation metrics: 
\begin{align*} 
\text{precision} &= \frac{|\widehat{\mathbb Y}_a \cap \sY_a| + |\widehat{\mathbb Y}_\emptyset \cap \sY_\emptyset|}{|\widehat{\mathbb Y}_a|+|\widehat{\mathbb Y}_\emptyset|}, \\ 
\text{recall} &= \frac{|\widehat{\mathbb Y}_a \cap \sY_a| + |\widehat{\mathbb Y}_\emptyset \cap \sY_\emptyset|}{|\sY_a|+|\sY_\emptyset|},
\end{align*}
where $\widehat{\mathbb Y}_a$ and $\sY_a$ are sets of word-by-word alignment pairs in prediction and ground-truth, respectively, e.g., $(s_i, t_j) \in \sY_a$ if $s_i$ aligns to $t_j$. 
The sets of $\widehat{\mathbb Y}_\emptyset$ and $\sY_\emptyset$ correspond to those of null-alignment that regards a word aligning to a null word $w_\emptyset$, e.g., $(s_k, w_\emptyset) \in \sY_\emptyset$ if $s_k$ is null alignment. 
Hence, $|\sY_\emptyset|$ is equal to the number of null alignment words in the ground-truth. 
Compared to previous metrics that only consider $\sY_a$ and $\widehat{\mathbb Y}_a$ \cite{maccartney:08}, ours consider null alignment equally as word-by-word alignment. 

\section{Background: Optimal Transport}
OT seeks the most efficient way to move mass from one measure to another. 
Remarkably, the OT problem induces the OT \emph{mapping} that indicates correspondences between the samples. 
While OT is frequently used as a distance metric between two measures, OT mapping is often the primary concern in alignment problems. 

%
Formally, the inputs to the optimal transport problem are a cost function and a pair of measures. 
On the premise of its application to monolingual word alignment, the following explanation assumes that the source and target sentences $s$ and $t$ and their word embeddings are at hand (see \Secref{sec:problem_definition}).
A \emph{cost} means a dissimilarity between $\vs_i$ and $\vt_j$ (source and target words) computed by a distance metric $c\colon \R^d \times \R^d \rightarrow \R_+$, such as Euclidean and cosine distances. 
The cost matrix $\mC \in \R_+^{n \times m}$ summarises the costs of any word pairs, that is, $\emC_{i,j}=c(\vs_i, \vt_j)$. 
A \emph{measure} means a weight each word has.
The concept of measure corresponds to \emph{fertility} introduced in IBM Model $3$ \cite{brown-etal-1993-mathematics}, which defines how many target (source) words a source (target) word can align. 
In summary, the mass of words in $s$ and $t$ is represented as arbitrary measures $\va \in \R_+^n$ and $\vb \in \R_+^m$, respectively.%
\footnote{Subsequently, we use the terms \emph{mass} and \emph{fertility} interchangeably to indicate each element of measures.} 
%
Finally, the OT problem identifies an alignment matrix $\mP$ with which the sum of alignment costs is minimised under the cost matrix $\mC$:
\begin{equation}
    L_C(\va, \vb) \coloneqq \min_{\mP \in \mU(\va, \vb)} \innerproduct{\mC}{\mP},
\end{equation}
where $\innerproduct{\mC}{\mP} \coloneqq \sum_{i, j} \emC_{i,j} \emP_{i,j}$. 
The alignment matrix $\mP$ belongs to $\mU(\va, \vb) \subseteq \R_+^{n \times m}$ that is a set of valid alignment matrices,
introduced in the following sections. 
With this formulation, 
we can seek
the most plausible word alignment matrix $\mP$. 

\subsection{Balanced Optimal Transport}
BOT \cite{kantorovich-1942} assumes that $\va$ and $\vb$ are probability distributions, i.e., $\va \in \Sigma_n$ and $\vb \in \Sigma_m$, respectively, where $\Sigma_n$ is the probability simplex: $\Sigma_n \coloneqq \{\vp \in \R_+^n: \sum_i \evp_i =1\}$. 
In this case, alignment matrices must live in the following space:
\begin{equation}
\resizebox{\linewidth}{!}{$
    \mU^b(\va, \vb) \! \coloneqq \! \{\mP \in \R_+^{n \times m} \!: \! \mP\identity_m=\va, \mP^\top\identity_n=\vb\},
    $}
\end{equation}
where $\identity_n$ is the all-ones vector of size $n$.  
Under this constraint set, the BOT problem is a linear programming (LP) problem and can be solved by standard LP solvers \cite{Pele2009ICCV}. 
However, the non-differentiable nature makes it challenging to integrate into neural models.  

\paragraph{Regularised BOT}
The entropy-regularised optimal transport \cite{cuturi-2013}, initially aimed at improving the computational speed of BOT, is a differentiable alternative and thus can be directly integrated into neural models. 
The regularisation makes the objective as follows:
\begin{equation}
    L_C^\varepsilon(\va, \vb) \coloneqq \min_{\mP \in \mU^b(\va, \vb)} \innerproduct{\mC}{\mP} + \varepsilon H(\mP),
\end{equation}
where the function $H$ is the negative entropy of alignment matrix $\mP$, 
and $\varepsilon$ controls the strength of the regularisation; with sufficiently small $\varepsilon$, $L_C^\varepsilon$ well approximates the exact BOT. 
The optimisation problem can be efficiently solved using the Sinkhorn algorithm \cite{sinkhorn}.  

\subsection{Relaxation of BOT Constraint}
Despite its success, the hard constraint $\mU^b$ of the BOT that aligns all words often makes it sub-optimal for some alignment problems where null alignment is more or less pervasive, such as unbalanced word alignment. 
POT and UOT introduced subsequently relax this constraint to allow certain words to be left unaligned.  

\paragraph{Partial Optimal Transport}
POT \cite{caffarelli-et-al-2010,figalli-2010} relaxes BOT by aligning only a fraction $m$ of the fertility, with the following constraint set in place of $\mU^b$: 
\begin{align}
\mU^p(\va, \vb) & \! \coloneqq \! \{\mP \in \R_+^{n \times m} \!: \! \mP\identity_m \leq \va, \mP^\top\identity_n \leq \vb, \\[-10pt]
& \identity_m^\top\mP^\top\identity_n = m\}. \label{eq:pot}
\end{align}
The fraction $m$ is bound as $m \leq \min (\| \va \|_1, \| \vb \|_1)$ where $\| \cdot \|_1$ represents the $L^1$ norm. 
While POT can be solved by standard LP solvers, it can also be regularised as in the BOT and solved with the Sinkhorn algorithm \cite{benamou-et-al-2015}. 

\paragraph{Unbalanced Optimal Transport}
UOT relaxes BOT by introducing soft constraints that penalise marginal deviation. 
\begin{align}
    L_C^\varepsilon(\va, \vb) & \coloneqq \min_{\mP \in \R_+^{n \times m}} \innerproduct{\mC}{\mP} + \varepsilon H(\mP) \\
    & + \tau_a D(\mP\identity_m, \va) + \tau_b D(\mP^\top\identity_n, \vb),
\end{align}
where $D$ is a divergence and $\tau_a$ and $\tau_b$ control how much mass deviations are penalised. 
Notice that the unbalanced formulation seeks alignment matrices in the entire $\R_+^{n \times m}$. 
In this study, we adopt the Kullback--Leibler divergence as $D$ because of its simplicity in computation \cite{cortes-et-al-2015,chizat-et-al-2016}. 

\section{Proposal: Unbalanced Word Alignment as Optimal Transport Problem}
In this study, we aim at formulating monolingual word alignment with high null frequencies in a natural way. 
For this purpose, we leverage OT with different constraints and reveal their features on this problem. 
We adopt the basic and generic cost matrices and measures instead of engineering them to avoid obfuscating the difference between BOT, POT, and UOT.

\subsection{Cost Function and Measures}
\label{sec:definition_cost_measure}
We obtain contextualised word embeddings from a pre-trained language model as adopted in the previous word alignment methods \cite{simalign,dou-neubig-2021-word}. 
Specifically, we concatenate source and target sentences and input to the language model to obtain the $\ell$-th layer hidden outputs. 
We then compute a word embedding by mean pooling the hidden outputs of its subwords. 

As a cost function, we use the cosine and Euclidean distances\footnote{Although the cosine distance is not a proper metric due to its in-satisfaction of the triangle inequality, we adopt it for its prevalence and empirical efficacy in NLP.} to obtain the cost matrix $\mC$, which are the most commonly used semantic dissimilarity metrics in NLP. 
In addition, we employ the distortion introduced in IBM Model $2$ \cite{brown-etal-1993-mathematics} when computing a cost to discourage aligning words appearing in distant positions. 
\citet{simalign} modelled the distortion between $s_i$ and $t_j$ as $\kappa\left(i/n-j/m\right)^2$, where $\kappa$ scales the value to range in $[0, \kappa]$. 
The value becomes larger if the relative positions of $i/n$ and $j/m$ are largely different. 
Each entry of the cost matrix is then modulated by the corresponding distortion value.
Note that the cost matrix is scaled in its computation process by using the min-max normalisation to make sure all entries lie in $[0, 1]$.\footnote{The scaling is not mathematically necessary; however, it is crucial for numerical stability in solving OT problems with the Sinkhorn algorithm; without the scaling, parameter tuning of $\varepsilon$ becomes challenging as the relative scale of the alignment cost $\innerproduct{\mC}{\mP}$ may deviate a lot.} 

Fertility determines the likelihood of words having alignment links. 
We use two standard measures adopted in previous studies that used OT in NLP tasks as discussed in \Secref{sec:related_work_ot4nlp}: the uniform distribution \cite{wmd} and $L^2$-norms of word embeddings \cite{yokoi-etal-2020-word}. 
On POT and UOT, we directly use these measures without scaling, for which $\emP_{i,j}$ may have an arbitrary positive value. 
We scale $\mP$ by the min-max normalisation so that we can handle alignment matrices of BOT, POT, and UOT by a unified manner. 

\subsection{Heuristics for Sparse Alignment}
While the Sinkhorn algorithm is a powerful tool for solving OT problems, one drawback is that a resultant alignment matrix becomes dense, i.e., each element has a non-zero weight. 
It is not straightforward to interpret a dense solution as an alignment matrix and thus dense matrices are better avoided.
As an empirical remedy, simple heuristics have been commonly used to make alignment matrices sparse: assuming top-$k$ elements based on their mass \cite{lee-etal-2022-toward,yu-etal-2022-optimal} or elements whose mass are larger than a threshold \cite{swanson-etal-2020-rationalizing,dou-neubig-2021-word} are aligned. 

We take the latter approach to avoid introducing an arbitrary assumption on fertility, i.e., the number of alignment links that a word can have. 
Specifically, we derive the final alignment matrix $\hat{\mP}$ using a threshold $\lambda$:
\begin{equation}
    \hat{\emP}_{i,j}=
    \begin{cases}
        \emP_{i,j}   &   \emP_{i,j} > \lambda,  \\
        0            &   \text{otherwise}.
    \end{cases}
\end{equation}
Our experiments reveal that this simple `patch' to obtain a sparse alignment can produce \emph{unbalanced} alignment, rather than just sparse, as we see in \Secref{sec:unsupervised_results}. 

\subsection{Application to Unsupervised Alignment}
We obtain contextualised word embeddings from a pre-trained masked language model without fine-tuning in the unsupervised setting. 
Such word embeddings are known to show relatively high cosine similarity between any random words \cite{ethayarajh-2019-contextual}, which blurs the actual similarity of semantically corresponding words. 
\citet{chen-etal-2020-improving-text} alleviated this phenomenon with a simple technique of centring the word embedding distribution. 
We apply the corpus mean centring that subtracts the mean word vector of the entire corpus from each word embedding. 

\subsection{Application to Supervised Alignment}
In the supervised alignment, we adopt linear metric learning \cite{gao-et-al-NIPS2016} that learns a matrix $\mW \in \R^{d \times d}$ defining a generalised distance between two embeddings: $c(\vs_i, \vt_j)=c(\mW\vs_i, \mW\vt_j)$. 
We train the entire model to learn parameters of $\mW$ and the pre-trained language model by minimising the binary cross-entropy loss: 
\begin{equation}
\resizebox{\linewidth}{!}{$
    L(\emP_{i,j}, \emY_{i,j})= -\emY_{i,j} \log \emP_{i,j} - (1-\emY_{i,j})\log (1-\emP_{i,j}),
    $}
\end{equation}
where $\mP$ and $\mY$ are the predicted and ground-truth alignment matrices, respectively. 
Specifically, $\emY_{i,j} \in \{0, 1\}$ indicates the ground-truth alignment between $s_i$ and $t_j$; $1$ means that alignment exists while $0$ means no alignment. 

\section{Experiment Settings}
We empirically investigate the characteristics of BOT, POT, and UOT on word alignment in both unsupervised and supervised settings. 
We refer to our OT-based alignment methods as OTAlign, hereafter. 
To alleviate performance variations due to neural model initialisation, all the experiments were conducted for $5$ times with different random seeds, and the means of the scores are reported. 

\paragraph{Dataset}
\begin{table}[t!]
\centering
\begin{tabular}{llrrr}
\toprule
              & & Train & Val & Test          \\ \midrule
\multicolumn{2}{l}{MSR-RTE}       & $600$   & $200$ & $800$ \\
\multicolumn{2}{l}{Edinburgh$++$} & $514$   & $200$ & $306$ \\
\multirow{4}{*}{MultiMWA}
 & MTRef         & $2,398$  & $800$ & $800$ \\
 & Wiki          & $2,514$  & $533$ & $1,052$ \\
 & Newsela       & --    & --  & $500$ \\
 & ArXiv         & --    & --  & $200$ \\ \bottomrule
\end{tabular}%
\caption{Statistics of evaluation datasets}
\label{tab:corpus_stats}
\end{table}

As datasets that provide human-annotated word alignment, we used Microsoft Research Recognizing Textual Entailment (\textbf{MSR-RTE})~\cite{msr-rte}, \textbf{Edinburgh$++$}~\cite{edinburgh++}, and Multi-Genre Monolingual Word Alignment (\textbf{MultiMWA})~\cite{lan-etal-2021-neural} as shown in \Tref{tab:corpus_stats}. 
MultiMWA consists of four subsets according to the sources of sentence pairs: \textbf{MTRef}, \textbf{Newsela}, \textbf{ArXiv}, and \textbf{Wiki}. 
Among them, Newsela and ArXiv are intended for a transfer-learning experiment, on which models trained using MTRef should be tested \cite{lan-etal-2021-neural}.
MSR-RTE and Edinburgh$++$ do not have an official split for validation. 
Hence, we subsampled a validation set from the training split, which was excluded from there. 
As the tradition of word alignment, there are two types of alignment links: sure and possible. 
The former indicates that an alignment was agreed upon by multiple annotators and thus preserves high confidence.  
Experiments were conducted in both `Sure and Possible' and `Sure Only' settings. 
For more details, see \Appendixref{sec:appendix:corpora}.

\paragraph{Pre-trained Language Model}
OTAlign as well as the recent powerful methods (\Secref{sec:related_worK_wa}) use contextualised word embeddings obtained from pre-trained (masked) language models. 
As the basic and standard model, we used BERT-base-uncased \cite{bert} for all the methods compared to directly observe the capabilities of different alignment mechanisms and to exclude performance differences owing to the underlying pre-trained models.\footnote{We indeed investigated the performance when adopting SpanBERT \cite{joshi-etal-2020-spanbert} as a pre-trained model. In summary, the results indicated that the underlying pre-trained model does not affect the overall trends of alignment methods, while their performance was consistently improved by a few percent owing to the better phrase representation.} 
For unsupervised word alignment, we used the $10$th layer in BERT that performs strongly in unsupervised textual similarity estimation \cite{bertscore}. 
For supervised alignment, we used the last hidden layer following the convention. 

\paragraph{OTAlign}
Our OT-based alignment methods require a cost function and fertility. 
As discussed in \Secref{sec:definition_cost_measure}, we experiment with \textbf{cosine} and \textbf{Euclidean} distances as cost functions and \textbf{uniform} distribution and $L^2$-\textbf{norm} of word embeddings as the fertility. 
Due to the space limitation, the main paper only discusses the results of a cost function and fertility that performed consistently strongly on the validation sets. 
\Appendixref{sec:appendix:more_results} draws the complete picture of different distance metrics and fertility and analyses their effects. 
We fixed the regularisation penalty $\varepsilon$ to $0.1$ throughout all experiments. 
The threshold $\lambda$ to sparsify alignment matrices was searched in $[0, 1]$ with $0.01$ interval to maximise the validation F$1$. 
More details of the implementation and experiment environment are described in \Appendixref{sec:appendix:implementation}.

\section{Unsupervised Word Alignment}
\begin{table*}[t!]
\centering
\resizebox{\textwidth}{!}{%
\begin{tabular}{@{}ccccccccccccccc@{}}
\toprule
\multicolumn{4}{c}{Dataset (sparse $\leftrightarrow$ dense)}      & \multicolumn{2}{c}{MSR-RTE}                 & \multicolumn{2}{c}{Newsela}                 & \multicolumn{2}{c}{EDB$++$}                 & \multicolumn{2}{c}{MTRef} & \multicolumn{2}{c}{Arxiv} & Wiki       \\ \midrule
\multicolumn{4}{c}{Alignment links}                              & S                    & S + P                & S                    & S + P                & S                    & S + P                & S           & S + P       & S           & S + P       & S          \\
\multicolumn{4}{c}{Null alignment rate (\%)}                               & $63.8$                 & $59.0$                 & $33.3$                 & $23.5$                 & $27.4$                 & $19.0$                 & $18.7$        & $11.2$        & $12.8$        & $12.2$        & $8.3$        \\ \midrule
\multicolumn{4}{c}{fast-align \cite{fastalign}} & $42.3$                 & $41.6$                 & $58.4$                 & $56.5$                 & $59.6$                 & $60.8$                 & $58.1$        & $58.0$        & $80.5$        & $80.5$        & $87.2$       \\
\multicolumn{4}{c}{SimAlign \cite{simalign}}    & $85.4$                 & $81.5$                 & $76.7$                 & $77.3$                 & $74.7$                 & $78.9$                 & $74.8$        & $75.8$        & $\underline{91.7}$  & $\underline{91.9}$  & $\underline{94.8}$ \\ \midrule
Type                       & Reg.     & cost       & mass        & \multicolumn{1}{l}{} & \multicolumn{1}{l}{} & \multicolumn{1}{l}{} & \multicolumn{1}{l}{} & \multicolumn{1}{l}{} & \multicolumn{1}{l}{} &             &             &             &             &            \\ \midrule
\multirow{2}{*}{BOT}        & --       & cosine     & uniform     & $20.6$                 & $22.5$                 & $41.4$                 & $46.9$                 & $49.0$                 & $55.0$                 & $50.4$        & $55.5$        & $65.6$        & $66.2$        & $66.5$       \\
                           & Sk       & cosine     & uniform     & $88.8$                 & $83.0$                 & $\underline{83.7}$                 & $\underline{79.4}$                 & $\underline{84.4}$           & $\underline{82.8}$           & $\underline{77.3}$  & $\underline{77.2}$  & $90.4$        & $\underline{90.9}$        & $\underline{93.9}$ \\
\multirow{2}{*}{POT}       & --       & cosine     & uniform     & $89.0$                 & $84.0$                 & $77.1$                 & $76.2$                 & $78.4$                 & $78.7$                 & $75.6$        & $\underline{76.2}$  & $84.3$        & $89.9$        & $\underline{94.5}$ \\
                           & Sk       & cosine     & uniform     & $\underline{92.2}$           & $\underline{86.4}$           & $\underline{84.6}$           & $\underline{79.8}$                 & $\underline{83.8}$                 & $\underline{82.3}$           & $\underline{77.0}$  & $\underline{76.6}$  & $\underline{91.5}$  & $90.3$        & $\underline{93.9}$ \\
\multirow{1}{*}{UOT}       & Sk       & cosine     & uniform     & $90.2$                 & $84.5$                 & $83.1$                 & $\underline{79.1}$                 & $\underline{84.7}$           & $\underline{82.5}$           & $\underline{77.2}$  & $\underline{77.1}$  & $90.0$        & $89.6$        & $\underline{93.8}$ \\ \bottomrule
\end{tabular}%
}
\caption{Unsupervised word alignment F$1$ scores (\%) measured on the test sets, where the \underline{underlined} scores are the best score and those within $1\%$ absolute differences. `S$+$P' and `S' are `Sure and Possible' and `Sure Only' settings, respectively. `Reg.' indicates the regulariser: `Sk' means the Sinkhorn.}
\label{tab:main_unsupervised_results_f1}
\end{table*}

\begin{figure*}[t!]
\centering
\begin{subfigure}[b]{0.47\textwidth}
         \centering
            \includegraphics[width=0.7\linewidth]{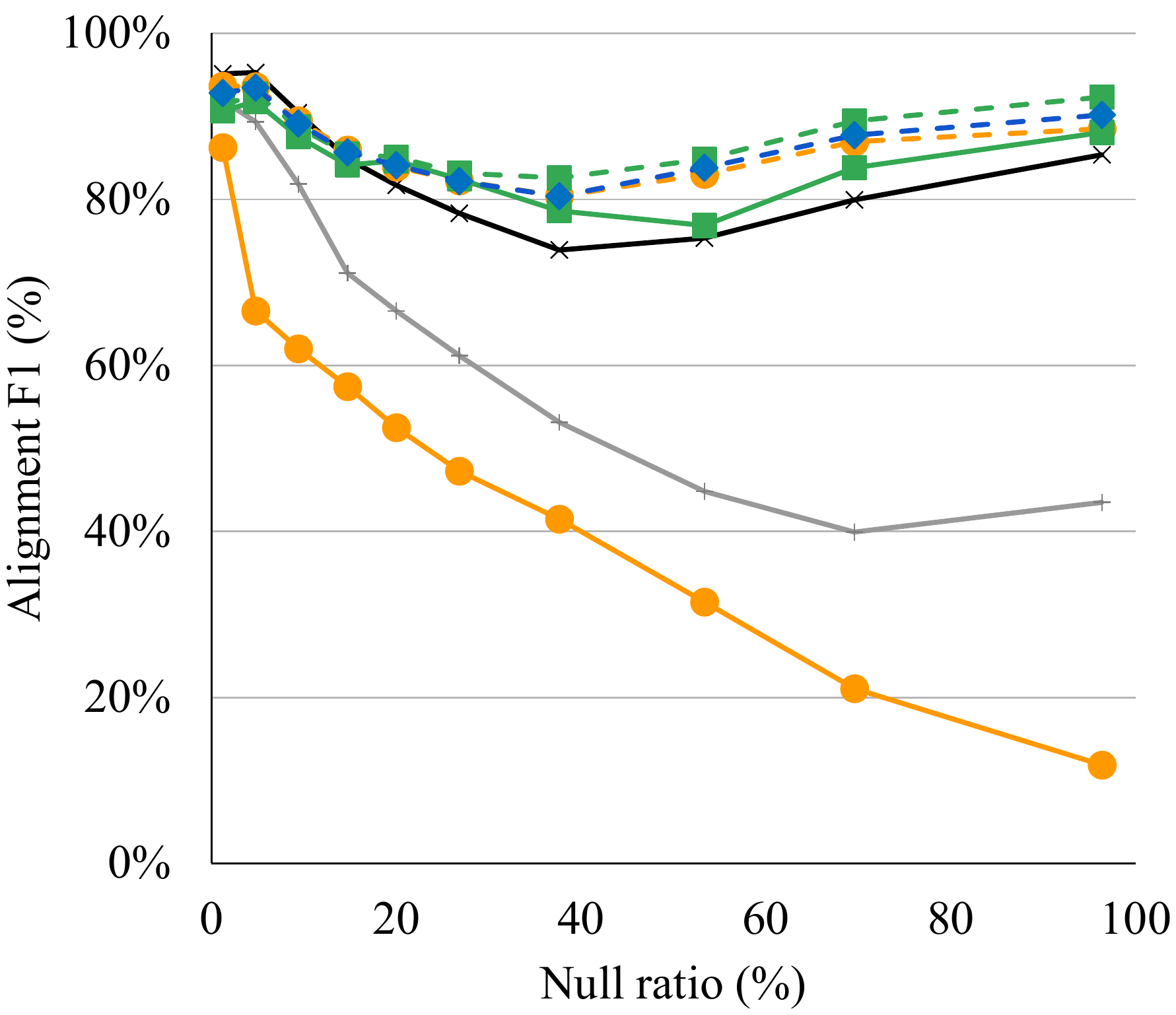}
            \caption{On all datasets}
            \label{fig:unsupervised_f1_by_null_rate:all}
     \end{subfigure}
     \hfill
    \begin{subfigure}[b]{0.47\textwidth}
         \centering
            \includegraphics[width=1.0\linewidth]{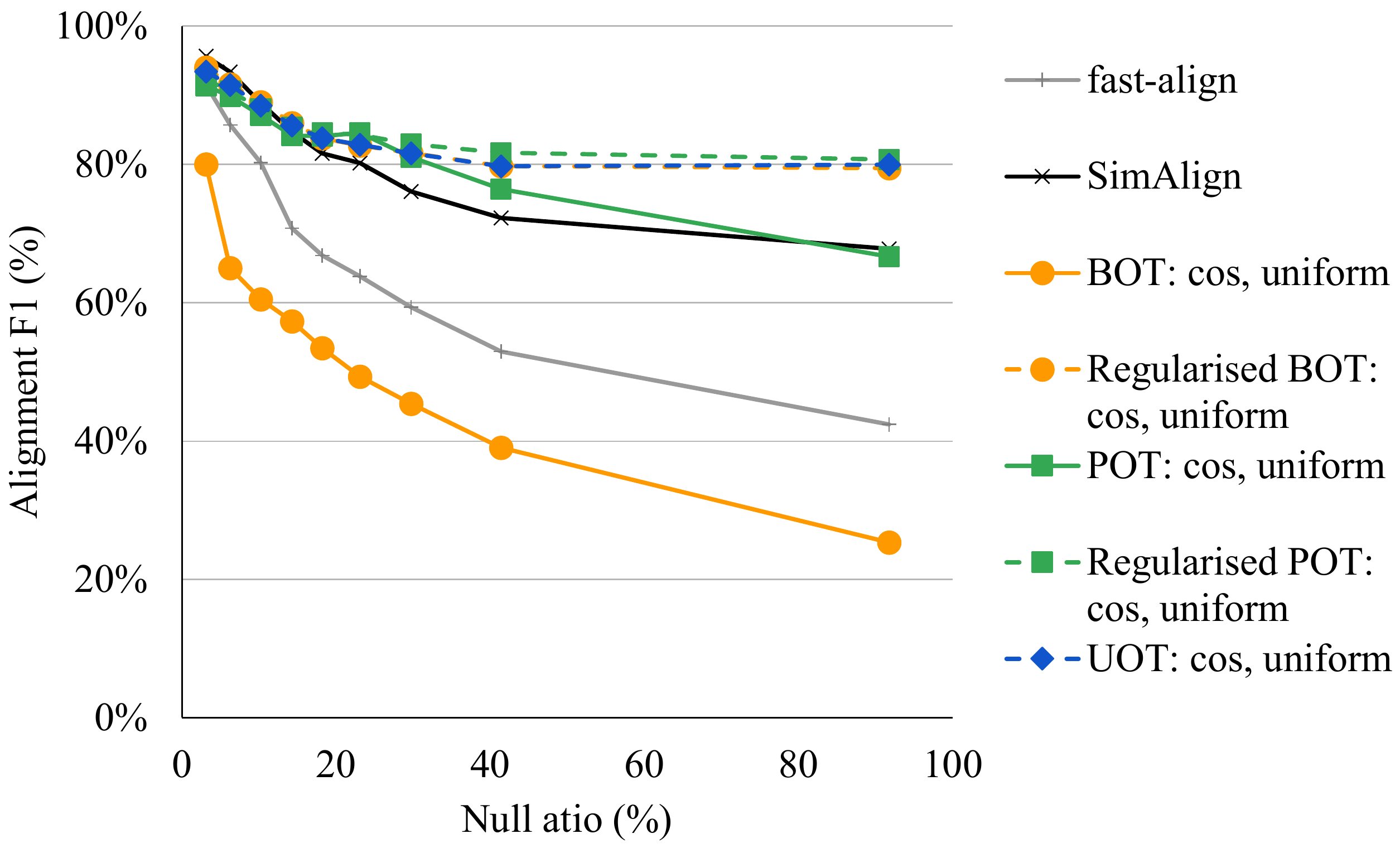}
            \caption{On datasets excluding MSR-RTE}
            \label{fig:unsupervised_f1_by_null_rate:wo_msr}
     \end{subfigure}
\caption{Unsupervised alignment F$1$ (\%) per null ratio}
\label{fig:unsupervised_f1_by_null_rate}
\end{figure*} 

We reveal features of OT-based methods on monolingual word alignment under unsupervised learning to segregate the effects of supervision on the pre-trained language model. 
Note that we use the validation sets for hyper-parameter tuning, which may not be strictly `unsupervised.'   
We still call this scenario `unsupervised' for convenience, believing that the preparation of small-scale annotation datasets should be feasible in most practical cases. 

\subsection{Settings}
As a conventional word alignment method, we compared OTAlign to the \textbf{fast-align} \cite{fastalign} that implemented IBM Model $2$. 
For fast-align, the training, validation, and test sets were concatenated and used to gather enough statistics for alignment. 
As the state-of-the-art unsupervised word alignment method, we compared OTAlign to \textbf{SimAlign} \cite{simalign}. 
While SimAlign was initially proposed for crosslingual word alignment, it is directly applicable to the monolingual setting. 
SimAlign computes a similarity matrix in the same manner as ours,\footnote{They use cosine \emph{similarity} instead of distance.} and aligns words using simple heuristics on the matrix. 
Specifically, we used the `IterMax' that performed best across many language pairs. 
IterMax conducts the `ArgMax' alignment iteratively (\citet{simalign} set the iteration as two times), which aligns a word pair if their similarity is the highest among each other. 
SimAlign has two hyper-parameters: one for the distortion ($\kappa$) and another for forcing null alignment if a word is not particularly similar to any target words. 
These values were tuned using the validation sets. 
 
For OTAlign, POT has a hyper-parameter $m$ to represent a fraction of fertility to be aligned, which we parameterise as $m=\widetilde{m} \min (\| \va \|_1, \| \vb \|_1)$. 
UOT has marginal deviation penalties of $\tau_a$ and $\tau_b$.\footnote{We set $\tau_a=\tau_b$ following the common trait to reduce the number of hyper-parameters \cite{chapel-et-al-nips2021}.} 
All of these hyper-parameters were searched in a range of $[0, 1]$ with $0.02$ interval to maximise the validation F$1$.  
For the distortion strength $\kappa$, we applied the same values tuned on SimAlign.

\subsection{Results: Primary Observations}
\label{sec:unsupervised_results}
\Tref{tab:main_unsupervised_results_f1} shows the F$1$ scores measured on the test sets.\footnote{The complete sets of results including precision and recall are in \Appendixref{sec:appendix:more_results}.}
We have the following observations:
\textbf{(i)~The best OT problem depends on null alignment ratios.} 
On datasets with higher null alignment ratios, i.e., Edinburgh$++$ and MTRef, regularised BOT, regularised POT, and UOT largely outperformed SimAlign. 
In particular, regularised POT performed the best on datasets with significantly high null alignment ratios, i.e., Newsela and MSR-RTE. 
On the other hand, when null alignment frequency is low, i.e., ArXiv and Wiki, regularised BOT, regularised POT, and UOT performed similarly. 
On these datasets, SimAlign also performed strongly thanks to the BERT representations.\footnote{\Tref{tab:unsupervised_results_f1} in \Appendixref{sec:appendix:more_results} shows that the simplest alignment by thresholding on a cosine similarity matrix using BERT embeddings outperforms SimAlign depending on datasets.} 
\textbf{(ii)~Thresholding on the alignment matrix makes it unbalanced.}
As expected, the unregularised BOT showed the worst performance because its constraint forces to align all words and prohibits null alignment.
In contrast, we observe that the performance of unregularised BOT and POT was significantly boosted by regularisation and thresholding on resultant alignment matrices (see the performance differences between with and without regulariser, denoted as `--' and `Sk', respectively, in \Tref{tab:main_unsupervised_results_f1}).

For further investigation, we binned the test samples across datasets (under the `Sure Only' setting) according to their null alignment ratios and observed the performance of the methods.\footnote{Each bin had roughly the same number of samples.} 
\Figref{fig:unsupervised_f1_by_null_rate}~(a) shows the trend of the F$1$ score on all the datasets according to the null alignment ratios. 
Although the F$1$ score of BOT naturally degrades as null alignment rates increase, the F$1$ score of regularised BOT stays closer to that of UOT owing to thresholding on the regularised solutions. 
Similarly, the regularised POT outperforms the unregularised POT. 
Therefore, we argue that thresholding is vital to obtain not only a sparse but also \emph{unbalanced} alignment matrix. 

\begin{table*}[t!]
\centering
\resizebox{\textwidth}{!}{%
\begin{tabular}{@{}cccccccccccccc@{}}
\toprule
\multicolumn{3}{c}{Dataset (sparse $\leftrightarrow$ dense)}             & \multicolumn{2}{c}{MSR-RTE}       & \multicolumn{2}{c}{Newsela}                         & \multicolumn{2}{c}{EDB$++$}                         & \multicolumn{2}{c}{MTRef}                   & \multicolumn{2}{c}{Arxiv}                   & Wiki                 \\ \midrule
\multicolumn{3}{c}{Alignment links}                                     & S          & S + P                & S                        & S + P                    & S                        & S + P                    & S                    & S + P                & S                    & S + P                & S                    \\
\multicolumn{3}{c}{Null alignment rate (\%)}                                      & $63.8$       & $59.0$                 & $33.3$                     & $23.5$                     & $27.4$                     & $19.0$                     & $18.7$                 & $11.2$                 & $12.8$                 & $12.2$                 & $8.3$                  \\ \midrule
\multicolumn{3}{c}{\cite{lan-etal-2021-neural}}        & $\underline{95.1}$ & $\underline{89.2}$           & $\underline{86.7}$               & $\underline{85.3}$               & $\underline{88.3}$               & $\underline{87.8}$               & $\underline{83.4}$           & $86.1$                 & $\underline{95.2}$           & $\underline{95.0}$           & $\underline{96.6}$           \\
\multicolumn{3}{c}{\cite{nagata-etal-2020-supervised}} & $\underline{95.0}$ & $\underline{89.2}$           & $79.4$                     & $82.4$                     & $86.9$                     & $\underline{87.2}$               & $\underline{82.9}$           & $\underline{88.0}$           & $89.1$                 & $89.5$                 & $\underline{96.5}$           \\ \midrule
Type                             & cost              & mass             &            & \multicolumn{1}{l}{} & \multicolumn{1}{l}{}     & \multicolumn{1}{l}{}     & \multicolumn{1}{l}{}     & \multicolumn{1}{l}{}     & \multicolumn{1}{l}{} & \multicolumn{1}{l}{} & \multicolumn{1}{l}{} & \multicolumn{1}{l}{} & \multicolumn{1}{l}{} \\ \midrule
\multirow{1}{*}{BOT}              & cosine            & norm             & $\underline{94.6}$ & $\underline{88.4}$           & $\underline{86.5}$               & $\underline{84.4}$               & \multicolumn{1}{l}{$85.7$} & \multicolumn{1}{l}{$85.4$} & $\underline{82.9}$           & $\underline{87.3}$           & $91.7$                 & $93.0$                 & $\underline{96.5}$           \\
\multirow{1}{*}{POT}             & cosine            & norm             & $\underline{94.6}$ & $\underline{88.4}$           & \multicolumn{1}{l}{$84.0$} & \multicolumn{1}{l}{$81.4$} & \multicolumn{1}{l}{$85.5$} & \multicolumn{1}{l}{$83.7$} & $82.0$                 & $85.2$                 & $93.0$                 & $92.2$                 & $95.5$                 \\
\multirow{1}{*}{UOT}             & cosine            & norm             & $\underline{94.8}$ & $\underline{89.0}$           & $\underline{86.8}$               & $\underline{84.7}$               & \multicolumn{1}{l}{$86.7$} & \multicolumn{1}{l}{$86.6$} & $\underline{82.9}$           & $\underline{87.4}$           & $92.5$                 & $92.8$                 & $\underline{96.7}$           \\ \bottomrule
\end{tabular}%
}
\caption{Supervised word alignment F$1$ scores (\%) measured on the test sets, where the \underline{underlined} scores are the best score and those within $1\%$ absolute differences.}
\label{tab:main_supervised_results_f1}
\end{table*}

Interestingly, most methods show a v-shape curve in \Figref{fig:unsupervised_f1_by_null_rate}~(a), i.e., the F$1$ score decreases first and then increases again. 
We identified this trend is due to the characteristics of MSR-RTE. 
Most sentence pairs with a high ($>45\%$) null alignment ratio come from MSR-RTE that consists of `hypothesis' and `text' of entailment pairs that tend to have largely different lengths. 
In these entailment pairs, most words in a (shorter) sentence would align with words in the (longer) pair, while the rest of the words in the pair have to be left unaligned. 
This characteristic is unique in MSR-RTE, which we conjecture affected the performance. 
In \Figref{fig:unsupervised_f1_by_null_rate}~(b), we removed MSR-RTE and drew trends of the alignment F$1$ score. 
The alignment F$1$ approximately becomes inversely proportional to the null alignment rate as expected.


\section{Supervised Word Alignment}
In this section, we evaluate the performance of OTAlign in supervised word alignment by comparing it to the state-of-the-art methods. 

\subsection{Settings}
\label{sec:experiments:supervised:settings}
We compared our OT-based alignment methods to the state-of-the-art supervised methods proposed by \citet{lan-etal-2021-neural} and \citet{nagata-etal-2020-supervised}. 
The method of \citet{lan-etal-2021-neural} uses a semi-Markov conditional random field combined with neural models to simultaneously conduct word and chunk alignment. 
Its high computational costs limit a chunk to be at most $3$-gram. 
The method proposed by \citet{nagata-etal-2020-supervised} models word alignment as span prediction in the same formulation as the SQuAD \cite{rajpurkar-etal-2016-squad} style question answering. 
These previous methods explicitly incorporate chunks in the alignment process. 
In contrast, our OT-based alignment is purely based on word-level distances represented as a cost matrix.

We trained alignment methods of the regularised BOT, regularised POT, and UOT in an end-to-end fashion using the Adafactor~\cite{adafactor} optimiser. 
The batch size was set to $64$ for datasets except for Wiki, for which $32$ was used due to longer sentence lengths. 
The training was stopped early, with $3$ patience epochs and a minimum delta of $\num{1.0e-4}$ based on the validation F$1$ score. 
Before evaluation, we searched the learning rate from $[\num{5.0e-5},\num{2.5e-4}]$ with a $\num{2.0e-5}$ interval using the validation sets. 
Other hyper-parameters were set to the same as the unsupervised settings. 

\subsection{Results}
\begin{figure*}[t!]
\centering
\begin{subfigure}[b]{0.47\textwidth}
         \centering
            \includegraphics[width=0.7\linewidth]{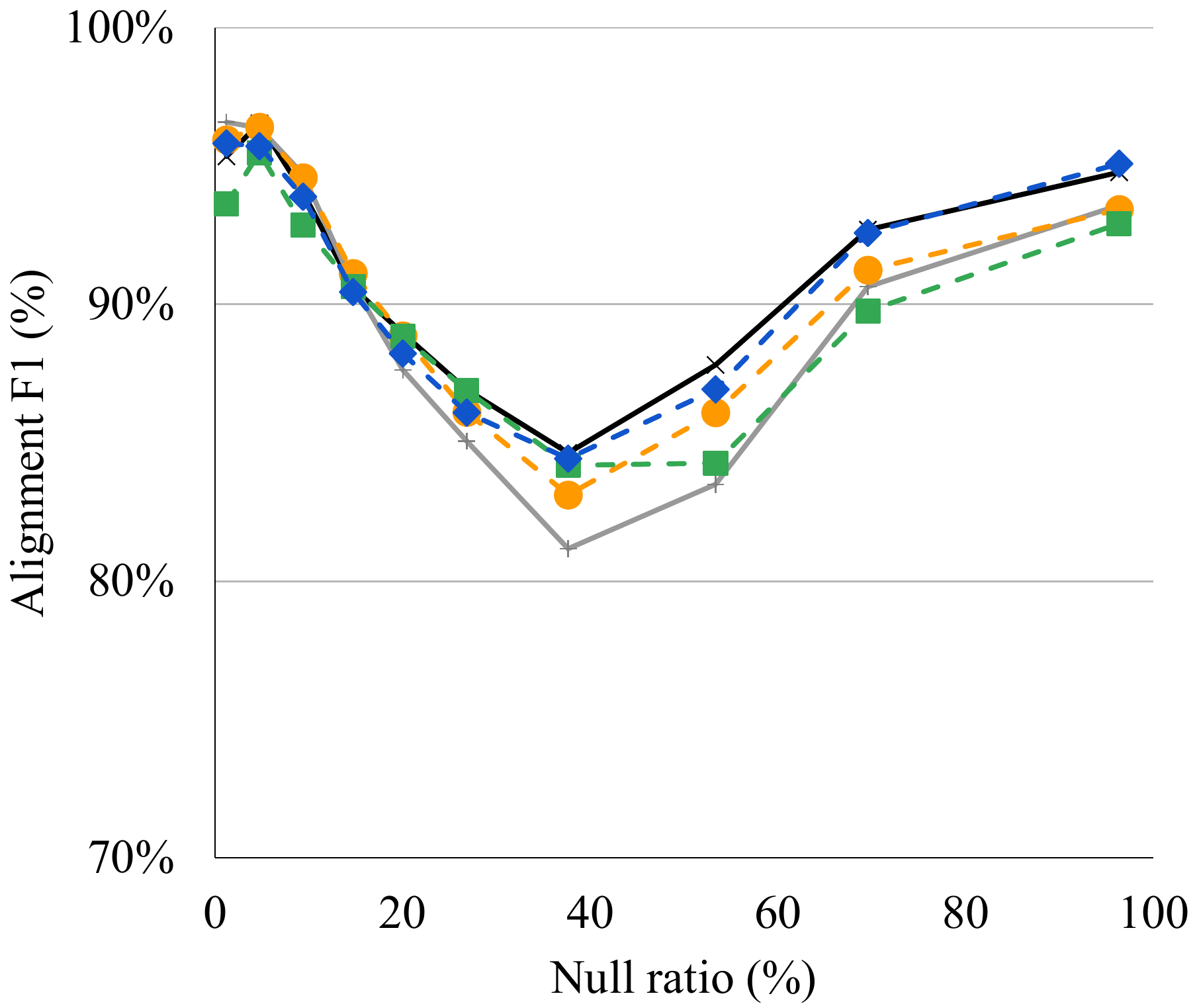}
            \caption{On all datasets}
            \label{fig:supervised_f1_by_null_rate:all}
     \end{subfigure}
     \hfill
    \begin{subfigure}[b]{0.47\textwidth}
         \centering
            \includegraphics[width=1.0\linewidth]{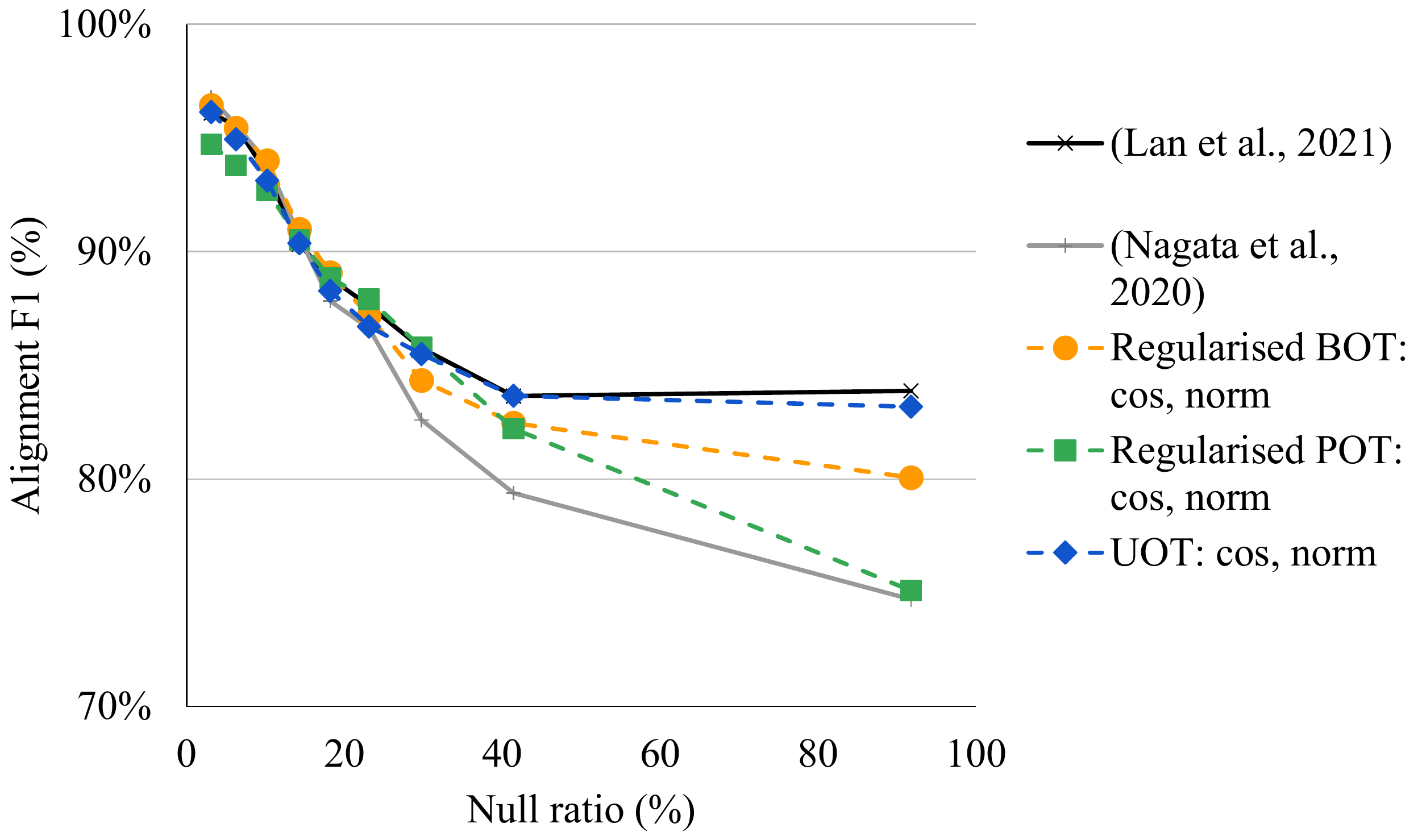}
            \caption{On datasets excluding MSR-RTE}
            \label{fig:supervised_f1_by_null_rate:wo_msr}
     \end{subfigure}
\caption{Supervised alignment F$1$ (\%) per null ratio}
\label{fig:supervised_f1_by_null_rate}
\end{figure*} 

\begin{figure*}[t!]
\centering
\begin{subfigure}[b]{0.3\textwidth}
         \centering
         \includegraphics[width=\textwidth]{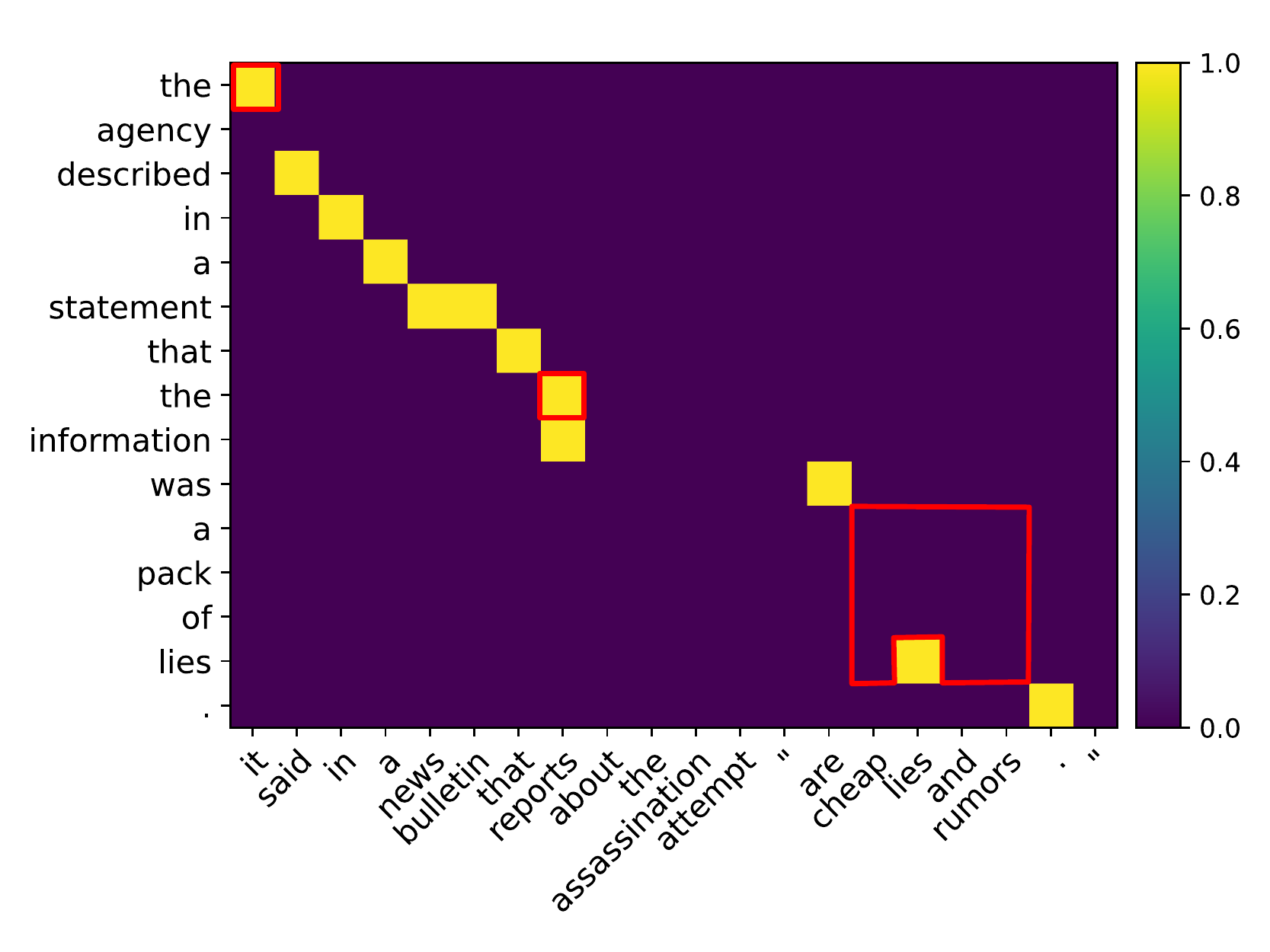}
         \caption{Alignment by \citet{lan-etal-2021-neural}}
     \end{subfigure}
     \hfill
\begin{subfigure}[b]{0.33\textwidth}
         \centering
         \includegraphics[width=\textwidth]{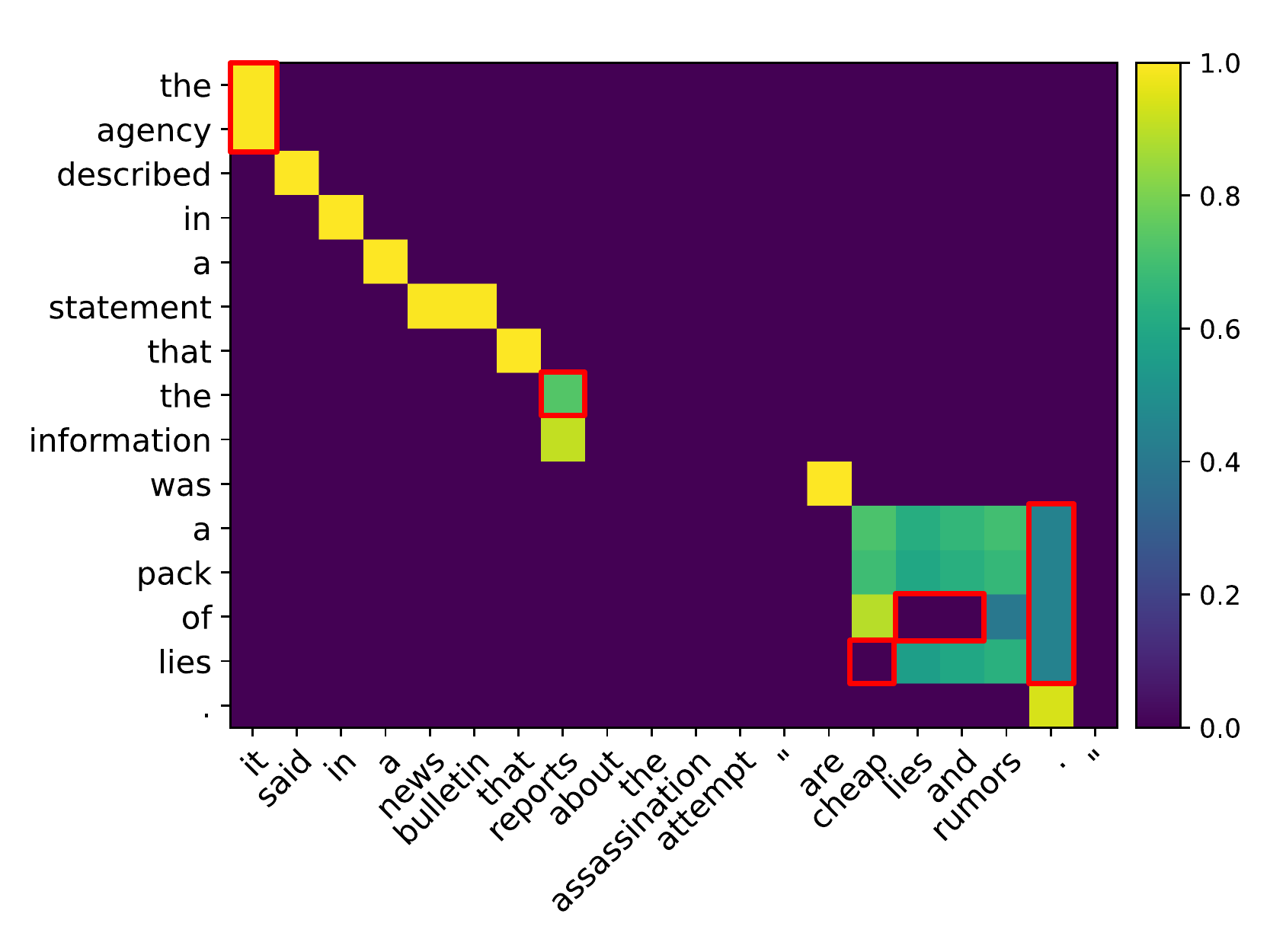}
         \caption{Alignment by \citet{nagata-etal-2020-supervised}}
     \end{subfigure}
     \hfill
\begin{subfigure}[b]{0.33\textwidth}
         \centering
         \includegraphics[width=\textwidth]{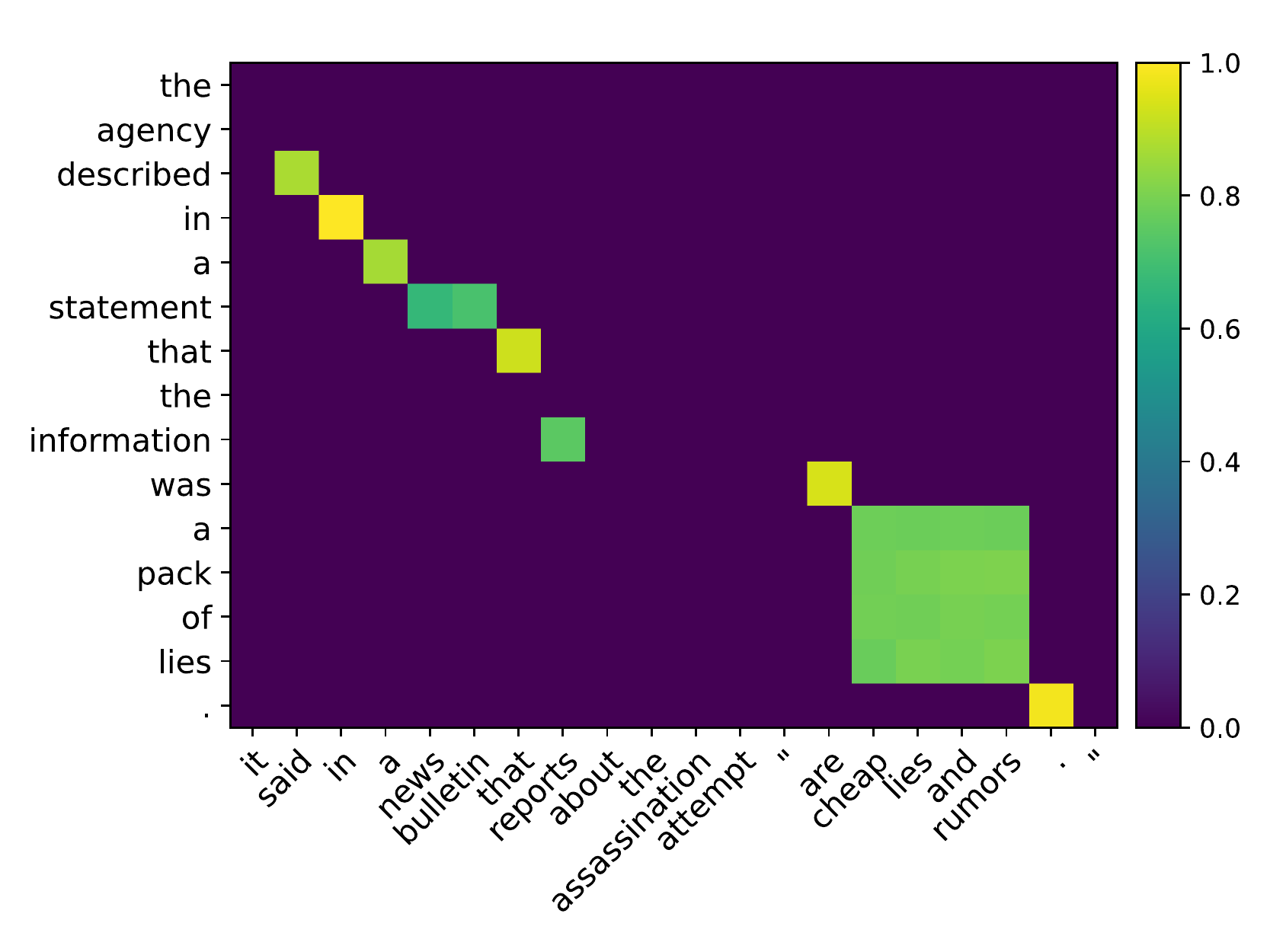}
         \caption{Alignment by supervised UOT}
     \end{subfigure}
\caption{Visualization of alignment matrix: \textcolor{red}{red} bounding boxes indicate alignment errors. UOT successfully identified many-to-many and null alignment.}
\label{fig:supervised_alignment}
\end{figure*} 

\Tref{tab:main_supervised_results_f1} shows the alignment F$1$ (\%) measured on the test sets.\footnote{The complete sets of results are available in \Appendixref{sec:appendix:more_results}.} 
The UOT-based alignment exhibits competitive performance against these state-of-the-art methods on the datasets with higher null alignment ratios. 
Notably, despite the simple and generic alignment mechanism based on UOT, it performed on par with \citet{lan-etal-2021-neural} who specially designed the model for monolingual word alignment. 
The UOT-based alignment also shows consistently higher alignment F$1$ scores compared to the regularised BOT and POT alignment. 
These results confirm that UOT well-captures the unbalanced word alignment problem. 
In addition, OT-based alignment showed better transferability as well as \citet{lan-etal-2021-neural} than \citet{nagata-etal-2020-supervised} as demonstrated on results of Newsela and Arxiv. 
We conjecture this is because our cost matrix has less inductive bias due to its simplicity. 
In contrast, \citet{nagata-etal-2020-supervised} directly learn to predict alignment spans. 

\Figref{fig:supervised_f1_by_null_rate} shows the trend of the F$1$ score according to the null alignment ratios assembled in the same way as \Figref{fig:unsupervised_f1_by_null_rate}. 
\Figref{fig:supervised_f1_by_null_rate}~(a) shows the trend on all datasets, demonstrating the robustness of OT-based methods against null alignment. 
OT-based alignment outperformed \citet{nagata-etal-2020-supervised} at sentences whose null alignment ratios are higher than $15\%$. 
The performances of the regularised BOT and UOT reverse around $20\%$ null alignment ratio; the regularised BOT outperformed UOT on a lower side ($0.7\%$ higher F$1$ on average except for the lowest edge) and UOT outperformed BOT on a higher side ($1.0\%$ higher F$1$ on average). 
Same with the unsupervised word alignment, all methods show the v-shape trend. 
\Figref{fig:supervised_f1_by_null_rate}~(b) shows the trend of the alignment F$1$ on the datasets except MSR-RTE. 
The performance of all methods becomes inversely proportional to the null alignment ratio as observed in the unsupervised case.

While UOT and previous methods are competitive on the datasets with high null alignment frequencies, we found that UOT has an advantage in longer phrase alignment. 
\Figref{fig:supervised_alignment}~(a), (b), and (c) visualise alignment matrices by \citet{lan-etal-2021-neural}, \citet{nagata-etal-2020-supervised}, and UOT, respectively, of the same sentence pair in \Figref{fig:gold_alignment}. 
The chunk size constraint in \citet{lan-etal-2021-neural}, $3$-gram at maximum, hindered the many-to-many alignment. 
\citet{nagata-etal-2020-supervised} also failed to complete this alignment because their method conducts one-to-many alignment bidirectionally and merges results. 
In contrast, UOT can align such a long phrase pair if the cost matrix is sufficiently reliable as demonstrated here.

\section{Summary and Future Work}
This study revealed the features of BOT, POT, and UOT on unbalanced word alignment and suggested that they perform sufficiently powerfully without tailor-made designs. 
In future, our OT-based methods can be enhanced for better phrase alignment by employing sophisticated phrase embedding models as discussed in Limitations section. 
We will also apply OT-based alignment to problems related yet with different constraints and objectives, e.g., crosslingual word alignment and text matching.

\section*{Limitations}
\begin{figure}[t!]
\centering
\includegraphics[width=1.0\linewidth]{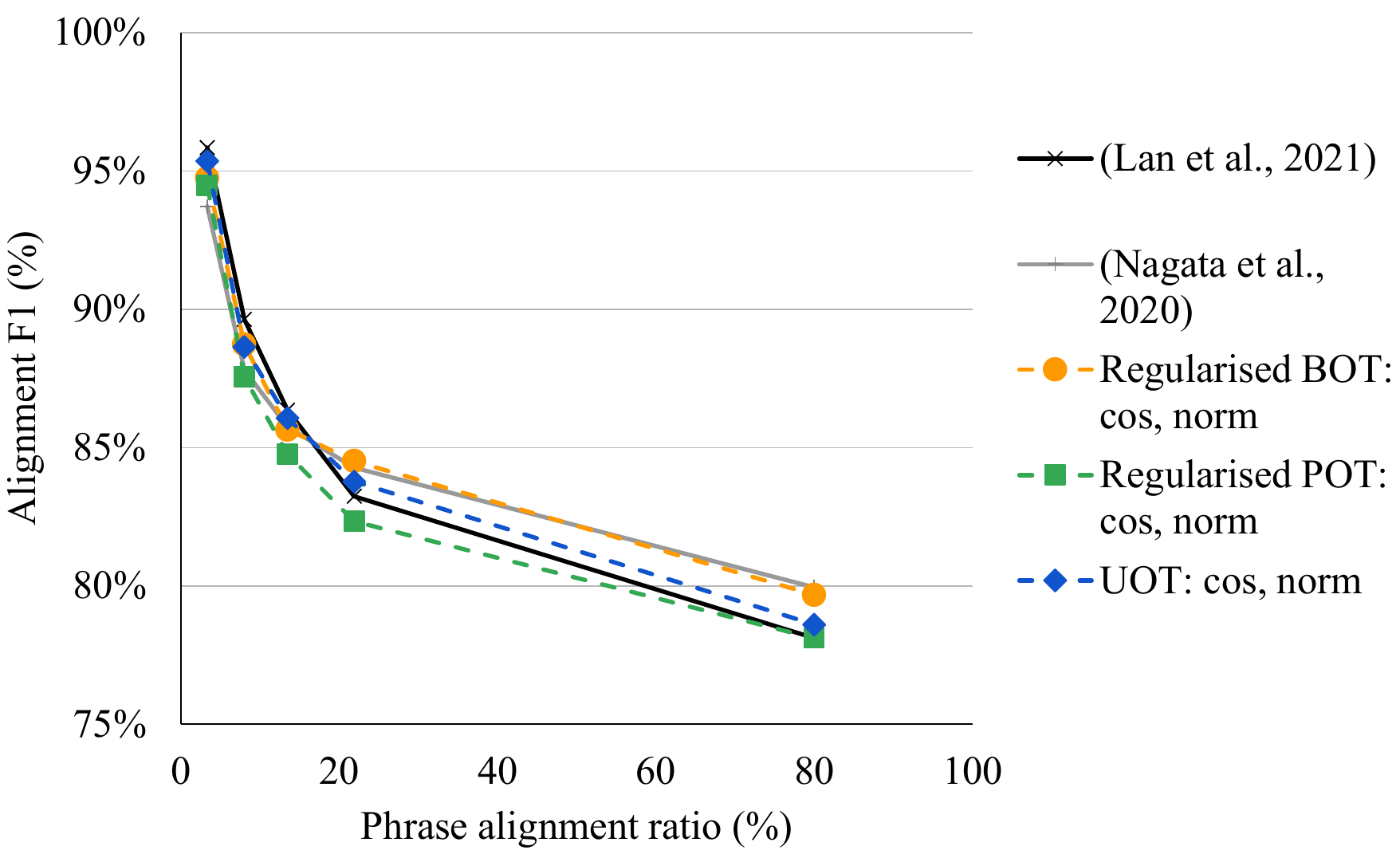}
\caption{Supervised alignment F$1$ (\%) per phrase alignment ratios}
\label{fig:supervised_f1_by_phrase_rate}
\end{figure} 

In this study, we used standard and basic word embeddings to highlight the characteristics of the different OT problems on unbalanced word alignment. 
This limits the capability of phrasal alignment. 
Similar to \Figref{fig:supervised_f1_by_null_rate}~(a), we binned all the test samples across datasets (under the `Sure Only' setting) according to their phrase alignment ratios and evaluated the performance of the supervised OT-based alignment methods.\footnote{In the evaluation datasets, phrase alignment is infrequent, which skewed the bin sizes: the first bin (less than $3.3\%$ phrase alignment ratio) size was $2.2$k and the rests were $\approx 350$.} 
Specifically, we regarded one-to-many, many-to-one, and many-to-many alignment as phrase alignment. 
\Figref{fig:supervised_f1_by_phrase_rate} shows the trend of the F$1$ score according to the phrase alignment ratios. 
Obviously, the F$1$ score degrades as more phrase alignment exists in a sentence pair. 

One of the straightforward ways to improve the phrase alignment is exploring pre-trained language models enhanced for span representations \cite{joshi-etal-2020-spanbert} and sophisticated methods for phrase representation composition \cite{yu-ettinger-2020-assessing}. 
In addition, phrase alignment can be addressed from the OT perspective, too, by conducting structure-aware \cite{alvarez-melis18a} and order-aware \cite{liu-et-al-2018-maching-natural} optimal transport. 
These directions constitute our future work.

\section*{Acknowledgements}
We sincerely appreciate the anonymous reviewers for their insightful comments and suggestions to improve the paper. 
This work was partially supported by JSPS KAKENHI Grant Number 22H03654.

\bibliography{paraphrase_alignment}
\bibliographystyle{acl_natbib}

\appendix

\section{Details of Evaluation Settings}
\subsection{Evaluation Datasets}
\label{sec:appendix:corpora}
MSR-RTE and Edinburgh$++$ have been commonly used for evaluating monolingual word alignment models \cite{yao-etal-2013-semi,sultan-etal-2014-back}. 
MultiMWA is the newest dataset that provides a larger number of examples with improved inter-annotator agreements. 

MSR-RTE annotated word alignment on sentence pairs for the $2006$ PASCAL RTE challenge \cite{dagan-et-al-2006,BarHaim2006TheSP}. 
Due to the nature of RTE pairs, the lengths of the source and target sentences tend to diverge, which results in a higher null-alignment ratio. 
In contrast, Edinburgh$++$ annotates word alignment on paraphrase pairs collected from machine translation (MT) references, novels, and news domains. 
MultiMWA annotated (or corrected the original) word alignment on (1) paraphrases extracted from multiple human references for MT evaluation \cite{Yao2014FeaturedrivenQA} (MTRef), (2) simple and complex sentence pairs sampled from Newsela-Auto \cite{jiang-etal-2020-neural} constructed for text simplification (Newsela), (3) aligned sentences extracted from arXiv\footnote{\url{https://arxiv.org/}} revision history (arXiv), and (4) aligned Wikipedia\footnote{\url{https://en.wikipedia.org/}} sentences based on the edit histories (Wiki).

\subsection{Implementation and Environment}
\label{sec:appendix:implementation}
All the BOT-, POT-, and UOT-based word alignment methods were implemented using PyTorch,\footnote{\url{https://pytorch.org/}} PyTorch Lightning,\footnote{\url{https://www.pytorchlightning.ai/}} Hugging Face Transformers \cite{wolf-etal-2020-transformers}, and Python Optimal Transport \cite{flamary2021pot} libraries. 

For the previous studies compared, we used the public implementations released by the authors with the necessary modifications. 
All the experiments were conducted on an NVIDIA Tesla V$100$ when GPU computation is needed. 

\section{Additional Experimental Results}
\label{sec:appendix:more_results}
Tables \ref{tab:unsupervised_results_f1} to \ref{tab:unsupervised_results_r} show the F$1$, precision, and recall scores of unsupervised word alignment with all the cost functions (cosine and Euclidean distances) and measures (the uniform distribution and $L^2$-norms). 
Tables \ref{tab:supervised_results_f1} to \ref{tab:supervised_results_r} indicate those of supervised word alignment experiments. 

\subsection{Effects of Cost Function and Measures}
\paragraph{Unsupervised Alignment}
The $7$th and $8$th rows from the top of \Tref{tab:unsupervised_results_f1} show the F$1$ scores of naive baselines that determine word alignment on the cost matrix with simple thresholding. 
That is, these baselines align words whose costs are smaller than the threshold. 
Their results indicate that cosine distance consistently outperforms Euclidean distance when word embeddings are naively used to estimate semantic similarity.  

The unregularised BOT and POT also prefer cosine distance as the cost function but their performances are more sensitive to measures, i.e., the uniform distribution outperformed the $L^2$-norms. 
The effect of the measures is most pronounced on the regularised BOT, likely because the uniform fertility makes the threshold selection simple. 
In contrast, UOT and regularised POT are less sensitive to the measures, likely because of their internal mechanism to allow null alignment.  

\paragraph{Supervised Alignment}
As shown in \Tref{tab:supervised_results_f1}, the UOT significantly improved over the unsupervised setting, outperforming the POT. 
In the supervised setting, UOT prefers the $L^2$-norms to model fertility than the uniform distribution. 
We conjecture that the BERT with linear metric learning successfully learns to adapt norms of word embeddings to well represent the fertility of words. 

\begin{table*}[t!]
\centering
\resizebox{\textwidth}{!}{%
\begin{tabular}{@{}ccccccccccccccc@{}}
\toprule
\multicolumn{4}{c}{Dataset (sparse $\leftrightarrow$ dense)}      & \multicolumn{2}{c}{MSR-RTE}                 & \multicolumn{2}{c}{Newsela}                 & \multicolumn{2}{c}{EDB$++$}                 & \multicolumn{2}{c}{MTRef} & \multicolumn{2}{c}{Arxiv} & Wiki       \\ \midrule
\multicolumn{4}{c}{Alignment links}                              & S                    & S + P                & S                    & S + P                & S                    & S + P                & S           & S + P       & S           & S + P       & S          \\
\multicolumn{4}{c}{Null alignment rate (\%)}                               & $63.8$                 & $59.0$                 & $33.3$                 & $23.5$                 & $27.4$                 & $19.0$                 & $18.7$        & $11.2$        & $12.8$        & $12.2$        & $8.3$        \\ \midrule
\multicolumn{4}{c}{fast-align \cite{fastalign}} & $42.3$                 & $41.6$                 & $58.4$                 & $56.5$                 & $59.6$                 & $60.8$                 & $58.1$        & $58.0$        & $80.5$        & $80.5$        & $87.2$       \\
\multicolumn{4}{c}{SimAlign \cite{simalign}}    & $85.4$                 & $81.5$                 & $76.7$                 & $77.3$                 & $74.7$                 & $78.9$                 & $74.8$        & $75.8$        & $\underline{91.7}$  & $\underline{91.9}$  & $\underline{94.8}$ \\ \midrule
Type                       & Reg.     & cost       & mass        & \multicolumn{1}{l}{} & \multicolumn{1}{l}{} & \multicolumn{1}{l}{} & \multicolumn{1}{l}{} & \multicolumn{1}{l}{} & \multicolumn{1}{l}{} &             &             &             &             &            \\
\bottomrule
\end{tabular}%
}
\caption{Unsupervised word alignment F$1$ scores (\%) measured on the test sets, where the \underline{underlined} scores are the best score and those within $1\%$ absolute differences. `S$+$P' and `S' are `Sure and Possible' and `Sure Only' settings, respectively. `Reg.' indicates the regulariser: `Sk' means the Sinkhorn. }
\label{tab:unsupervised_results_f1}
\end{table*}

\begin{table*}[t!]
\centering
\resizebox{\textwidth}{!}{%
\begin{tabular}{@{}ccccccccccccccc@{}}
\toprule
\multicolumn{4}{c}{Dataset (sparse $\leftrightarrow$ dense)}      & \multicolumn{2}{c}{MSR-RTE}                         & \multicolumn{2}{c}{Newsela}                         & \multicolumn{2}{c}{EDB$++$}                     & \multicolumn{2}{c}{MTRef}                           & \multicolumn{2}{c}{Arxiv}                           & Wiki                     \\ \midrule
\multicolumn{4}{c}{Alignment links}                              & S                        & S + P                    & S                        & S + P                    & S                    & S + P                    & S                        & S + P                    & S                        & S + P                    & S                        \\
\multicolumn{4}{c}{Null alignment rate (\%)}                               & $63.8$                     & $59.0$                     & $33.3$                     & $23.5$                     & $27.4$                 & $19.0$                     & $18.7$                     & $11.2$                     & $12.8$                     & $12.2$                     & $8.3$                      \\ \midrule
\multicolumn{4}{c}{fast-align \cite{fastalign}} & $43.3$                     & $43.0$                     & $59.8$                     & $57.0$                     & $60.8$                 & $60.9$                     & $57.9$                     & $57.2$                     & $79.7$                     & $79.6$                     & $86.5$                     \\
\multicolumn{4}{c}{SimAlign \cite{simalign}}    & $85.7$                     & $82.6$                     & $80.0$                     & $79.6$                     & $77.6$                 & $81.4$                     & $\underline{77.3}$               & $\underline{77.6}$               & $\underline{92.3}$               & $\underline{92.5}$               & $\underline{95.4}$               \\
\bottomrule
\end{tabular}%
}
\caption{Unsupervised word alignment precisions (\%) measured on the test sets, where the \underline{underlined} scores are the best and those within $1\%$ absolute differences. `S$+$P' and `S' are `Sure and Possible' and `Sure Only' settings, respectively. `Reg.' indicates the OT regulariser: `Sk' means the Sinkhorn.}
\label{tab:unsupervised_results_p}
\end{table*}

\begin{table*}[t!]
\centering
\resizebox{\textwidth}{!}{%
\begin{tabular}{@{}ccccccccccccccc@{}}
\toprule
\multicolumn{4}{c}{Dataset (sparse $\leftrightarrow$ dense)}      & \multicolumn{2}{c}{MSR-RTE}                 & \multicolumn{2}{c}{Newsela}                 & \multicolumn{2}{c}{EDB$++$}                 & \multicolumn{2}{c}{MTRef} & \multicolumn{2}{c}{Arxiv} & Wiki       \\ \midrule
\multicolumn{4}{c}{Alignment links}                              & S                    & S + P                & S                    & S + P                & S                    & S + P                & S           & S + P       & S           & S + P       & S          \\
\multicolumn{4}{c}{Null alignment rate (\%)}                               & $63.8$                 & $59.0$                 & $33.3$                 & $23.5$                 & $27.4$                 & $19.0$                 & $18.7$        & $11.2$        & $12.8$        & $12.2$        & $8.3$        \\ \midrule
\multicolumn{4}{c}{fast-align \cite{fastalign}} & $41.4$                 & $40.2$                 & $57.1$                 & $56.0$                 & $58.3$                 & $60.7$                 & $58.3$        & $58.7$        & $81.2$        & $81.3$        & $87.9$       \\
\multicolumn{4}{c}{SimAlign \cite{simalign}}    & $85.2$                 & $80.5$                 & $73.6$                 & $75.0$                 & $72.0$                 & $76.5$                 & $72.4$        & $74.1$        & $91.1$        & $91.4$        & $\underline{94.2}$ \\ \midrule
Type                       & Reg.     & cost       & mass        & \multicolumn{1}{l}{} & \multicolumn{1}{l}{} & \multicolumn{1}{l}{} & \multicolumn{1}{l}{} & \multicolumn{1}{l}{} & \multicolumn{1}{l}{} &             &             &             &             &            \\
\bottomrule
\end{tabular}%
}
\caption{Unsupervised word alignment recalls (\%) measured on the test sets, where the \underline{underlined} scores are the best and those within $1\%$ absolute differences. `S$+$P' and `S' are `Sure and Possible' and `Sure Only' settings, respectively. `Reg.' indicates the OT regulariser: `Sk' means the Sinkhorn.}
\label{tab:unsupervised_results_r}
\end{table*}

\begin{table*}[t!]
\centering
\resizebox{\textwidth}{!}{%
\begin{tabular}{@{}cccccccccccccc@{}}
\toprule
\multicolumn{3}{c}{Dataset (sparse $\leftrightarrow$ dense)}             & \multicolumn{2}{c}{MSR-RTE}       & \multicolumn{2}{c}{Newsela}                         & \multicolumn{2}{c}{EDB$++$}                         & \multicolumn{2}{c}{MTRef}                   & \multicolumn{2}{c}{Arxiv}                   & Wiki                 \\ \midrule
\multicolumn{3}{c}{Alignment links}                                     & S          & S + P                & S                        & S + P                    & S                        & S + P                    & S                    & S + P                & S                    & S + P                & S                    \\
\multicolumn{3}{c}{Null alignment rate (\%)}                                      & $63.8$       & $59.0$                 & $33.3$                     & $23.5$                     & $27.4$                     & $19.0$                     & $18.7$                 & $11.2$                 & $12.8$                 & $12.2$                 & $8.3$                  \\ \midrule
\multicolumn{3}{c}{\cite{lan-etal-2021-neural}}        & $\underline{95.1}$ & $\underline{89.2}$           & $\underline{86.7}$               & $\underline{85.3}$               & $\underline{88.3}$               & $\underline{87.8}$               & $\underline{83.4}$           & $86.1$                 & $\underline{95.2}$           & $\underline{95.0}$           & $\underline{96.6}$           \\
\multicolumn{3}{c}{\cite{nagata-etal-2020-supervised}} & $\underline{95.0}$ & $\underline{89.2}$           & $79.4$                     & $82.4$                     & $86.9$                     & $\underline{87.2}$               & $\underline{82.9}$           & $\underline{88.0}$           & $89.1$                 & $89.5$                 & $\underline{96.5}$           \\
\bottomrule
\end{tabular}%
}
\caption{Supervised word alignment F$1$ scores (\%) measured on the test sets, where the \underline{underlined} scores are the best score and those within $1\%$ absolute differences.}
\label{tab:supervised_results_f1}
\end{table*}

\begin{table*}[t!]
\centering
\resizebox{\textwidth}{!}{%
\begin{tabular}{@{}cccccccccccccc@{}}
\toprule
\multicolumn{3}{c}{Dataset (sparse $\leftrightarrow$ dense)}             & \multicolumn{2}{c}{MSR-RTE}           & \multicolumn{2}{c}{Newsela}                         & \multicolumn{2}{c}{EDB$++$}                         & \multicolumn{2}{c}{MTRef}                   & \multicolumn{2}{c}{Arxiv}                   & Wiki                 \\ \midrule
\multicolumn{3}{c}{Alignment links}                                     & S          & S + P                    & S                        & S + P                    & S                        & S + P                    & S                    & S + P                & S                    & S + P                & S                    \\
\multicolumn{3}{c}{Null alignment rate (\%)}                                      & $63.8$       & $59.0$                     & $33.3$                     & $23.5$                     & $27.4$                     & $19.0$                     & $18.7$                 & $11.2$                 & $12.8$                 & $12.2$                 & $8.3$                  \\ \midrule
\multicolumn{3}{c}{\cite{lan-etal-2021-neural}}        & $\underline{95.0}$ & $\underline{89.9}$               & $88.1$                     & $85.7$                     & $\underline{89.7}$               & $\underline{88.4}$               & $\underline{84.1}$           & $86.0$                 & $\underline{95.4}$           & $\underline{95.1}$           & $\underline{96.8}$           \\
\multicolumn{3}{c}{\cite{nagata-etal-2020-supervised}} & $\underline{95.0}$ & $\underline{90.2}$               & $80.3$                     & $82.2$                     & $88.5$                     & $\underline{88.1}$               & $83.3$                 & $\underline{87.4}$           & $87.8$                 & $88.0$                 & $\underline{96.7}$           \\
\bottomrule
\end{tabular}%
}
\caption{Supervised word alignment precision (\%) measured on the test sets, where the \underline{underlined} scores are the best and those within $1\%$ absolute differences.}
\label{tab:supervised_results_p}
\end{table*}

\begin{table*}[t!]
\centering
\resizebox{\textwidth}{!}{%
\begin{tabular}{@{}cccccccccccccc@{}}
\toprule
\multicolumn{3}{c}{Dataset (sparse $\leftrightarrow$ dense)}             & \multicolumn{2}{c}{MSR-RTE}       & \multicolumn{2}{c}{Newsela}                         & \multicolumn{2}{c}{EDB$++$}                         & \multicolumn{2}{c}{MTRef}                   & \multicolumn{2}{c}{Arxiv}                   & Wiki                 \\ \midrule
\multicolumn{3}{c}{Alignment links}                                     & S          & S + P                & S                        & S + P                    & S                        & S + P                    & S                    & S + P                & S                    & S + P                & S                    \\
\multicolumn{3}{c}{Null alignment rate (\%)}                                      & $63.8$       & $59.0$                 & $33.3$                     & $23.5$                     & $27.4$                     & $19.0$                     & $18.7$                 & $11.2$                 & $12.8$                 & $12.2$                 & $8.3$                  \\ \midrule
\multicolumn{3}{c}{\cite{lan-etal-2021-neural}}        & $\underline{95.2}$ & $\underline{88.5}$           & $\underline{85.4}$               & $\underline{84.9}$               & $\underline{87.0}$               & $\underline{87.1}$               & $\underline{82.6}$           & $86.2$                 & $\underline{95.0}$           & $\underline{94.9}$           & $\underline{96.4}$           \\
\multicolumn{3}{c}{\cite{nagata-etal-2020-supervised}} & $\underline{95.0}$ & $\underline{88.3}$           & $78.5$                     & $82.6$                     & $85.4$                     & $\underline{86.3}$               & $\underline{82.5}$           & $\underline{88.6}$           & $90.5$                 & $91.0$                 & $\underline{96.3}$           \\
\bottomrule
\end{tabular}%
}
\caption{Supervised word alignment recall (\%) measured on the test sets, where the \underline{underlined} scores are the best and those within $1\%$ absolute differences.}
\label{tab:supervised_results_r}
\end{table*}

\end{document}